\documentclass{article} 
\usepackage{iclr2027_conference,times}
\usepackage{amsmath}
\usepackage{amssymb}
\usepackage{graphicx}

\usepackage{amsmath,amsfonts,bm}

\def\eqref#1{equation~\ref{#1}}

\def\1{\bm{1}}

\DeclareMathAlphabet{\mathsfit}{\encodingdefault}{\sfdefault}{m}{sl}
\SetMathAlphabet{\mathsfit}{bold}{\encodingdefault}{\sfdefault}{bx}{n}

\usepackage{algorithm}
\usepackage{algpseudocode}
\usepackage{hyperref}
\usepackage{url}
\usepackage{booktabs}

\title{AutoQS: Automated Quantization Strategy Design with Large Language Models}

\author{
Zhen Liu$^{1}$\thanks{Equal contribution.}
\quad
Zhiqin Yang$^{2}$\footnotemark[1]
\quad
Yonggang Zhang$^{2}$ \\
Jianyi Liu$^{1}$
\quad
Jinjun Wang$^{1}$
\quad
Jingwen Fu$^{3}$\thanks{Corresponding author: \texttt{fujingwen@bza.edu.cn}.}
\\[0.5em]
$^{1}$Xi'an Jiaotong University \\
$^{2}$The Hong Kong University of Science and Technology \\
$^{3}$Zhongguancun Academy
}

\iclrfinalcopy  

\begin{document}
\maketitle

\lhead{}

\begin{abstract}
Quantization reduces LLM deployment costs, but effective strategies must balance model quality and storage.
LLM-driven search can automate strategy design through repeated proposal and evaluation.
This process must identify useful strategies within a limited evaluation budget.
The key challenge is that \textit{actual quantization is expensive, while inexpensive proxies may not reflect quantized-model performance.}
To address this challenge, we present AutoQS, an \textbf{Auto}mated \textbf{Q}uantization \textbf{S}trategy Design with LLMs.
AutoQS keeps weight precision fixed and searches group-size allocations across layers and operators under deployment constraints.
Evaluation history guides coordinated LLM proposals, while deterministic validation ensures compatible allocations.
An inexpensive quality proxy evaluates candidates, and an candidate repository retains alternative allocations to inform subsequent proposals.
Periodic actual quantization complements these proxy evaluations with measured quality and checkpoint storage, providing feedback
for further search within the evaluation budget.
Experiments on Qwen2.5-7B-Instruct show that AutoQS improves seven-task average accuracy over uniform-group GPTQ and AWQ by 0.70 and 1.12 percentage points, respectively, at four-bit weight precision.

\end{abstract}

\section{Introduction}
\label{sec:introduction}

Quantization reduces the storage and memory requirements of large language models (LLMs) by representing weights with fewer bits~\citep{frantar2023gptq,lin2024awq}.
Preserving model quality under this compression requires careful choices about how quantization is applied throughout the model.
Even at a fixed weight precision, quantization granularity remains an important design choice.
In group-wise quantization, groups of weights share quantization parameters, such as scales.
Smaller groups provide finer granularity at the cost of additional metadata.
Because model components respond differently to quantization, a uniform allocation need not provide the most effective use of this granularity~\citep{dong2019hawq,heo2024adadim}.
Designing a heterogeneous allocation therefore requires coordinating decisions across layers and operators while respecting the configurations supported by the quantization implementation.

Existing research provides two complementary foundations for automating this process.
Quantization methods improve how low-precision weights are obtained.
GPTQ, a post-training quantization method for generative pre-trained transformers, uses approximate second-order information to reduce reconstruction error~\citep{frantar2023gptq}.
Activation-aware Weight Quantization (AWQ) protects salient weight channels using activation statistics, while OmniQuant learns calibration parameters to improve low-bit reconstruction~\citep{lin2024awq,shao2024omniquant}.
Automated allocation methods, including Hardware-Aware Automated Quantization (HAQ), further demonstrate the value of using evaluation feedback to choose quantization configurations~\citep{wang2019haq}.
In parallel, LLM-driven systems such as FunSearch and Evolution of Heuristics automate program and heuristic design through repeated proposal and evaluation~\citep{romeraparedes2024funsearch,liu2024eoh}.
Recent systems further show how evaluation feedback can guide
subsequent LLM-generated changes during iterative
search~\citep{ye2024reevo,novikov2025alphaevolve}.
These developments motivate an analogous approach to
quantization strategy design, where proposed allocations
are evaluated and the resulting evidence informs later proposals.

Applying this approach to quantization introduces two connected difficulties.
First, a proposed strategy must become a valid allocation before it can be evaluated.
Changes across layers and operators must satisfy shared-configuration requirements, and differently expressed proposals may encode the same allocation.
Second, evaluating a valid allocation at full fidelity requires constructing and testing a quantized checkpoint.
Repeating this process for every proposal would consume the experimental budget quickly.
Inexpensive proxies support broader exploration, but their scores do not directly establish downstream accuracy or the size of a materialized checkpoint.
These difficulties raise a central question.
\textbf{\emph{How can LLM-driven search turn quantization proposals into executable strategies and improve them using a limited number of actual quantization evaluations?}}

We address this question with \textbf{AutoQS}, a framework for \textbf{Auto}mated \textbf{Q}uantization \textbf{S}trategy Design with LLMs.
AutoQS keeps weight precision fixed within each run and searches group-size allocations across layers and fusion-compatible operator groups.
To turn proposals into executable strategies, it combines \textbf{structured LLM edits with deterministic validation}.
The LLM receives a parent allocation, previous evaluation results, and a candidate repository, and proposes sparse edits that can jointly change several operator decisions.
Each proposal is checked for consistency with the parent and for compatibility with the supported quantization configurations.
Equivalent allocations are detected from their canonical representation, and deterministic novelty repair seeks an unseen valid alternative when an allocation is repeated.
This gives the LLM scope to propose coordinated changes while enforcing the concrete requirements of the quantization implementation.

Multi-fidelity optimization provides a precedent for combining
inexpensive approximations with costly target
evaluations~\citep{kandasamy2017multi}.
AutoQS adopts this broad evaluation principle through \textbf{low-cost
candidate assessment and periodic evaluation of quantized checkpoints}.
Valid, unseen candidates receive a scalar proxy score together with diagnostic statistics and a storage estimate.
Maintaining a population of evaluated candidates can provide
varied contexts for subsequent LLM proposals, as demonstrated
by FunSearch~\citep{romeraparedes2024funsearch}.
In AutoQS, the candidate repository supports parent selection and retains
alternative allocation patterns for later proposals.
However, proxy-based exploration alone cannot verify the quality of the resulting quantized models.
AutoQS therefore complements this loop with \textbf{periodic quantization and checkpoint feedback}.
At a fixed interval, it evaluates the latest eligible candidate in the current search window.
The selected strategy undergoes fresh GPTQ quantization~\citep{frantar2023gptq}, followed by development-task evaluation and checkpoint-size measurement.
The resulting observations inform later search decisions, connecting proposal generation to the behavior of actual quantized models.
The default protocol permits at most 20 such evaluation opportunities over 100 search iterations.
After candidate selection is frozen, a broader seven-task evaluation assesses the selected strategies without feeding its results back into search.

At four-bit weight precision, AutoQS improves seven-task average accuracy over uniform-group GPTQ and AWQ by 0.70 and 1.12 percentage points, respectively, on Qwen2.5-7B-Instruct.
Matched-seed ablations support the roles of LLM proposals and checkpoint feedback in improving search-time performance.
Case studies further examine how the discovered strategies allocate quantization granularity across model components and transfer to related models without additional allocation search.

Our contributions are threefold as follows:
\begin{itemize}
    \item \textbf{Automated quantization strategy design.}
    We introduce AutoQS to automate heterogeneous group-size allocation at fixed weight precision.
    The framework uses an LLM to propose coordinated strategy changes while enforcing operator compatibility and allocation validity.

    \item \textbf{Search with periodic checkpoint feedback.}
    We combine inexpensive candidate evaluation and candidate search history with a bounded schedule of actual quantization evaluations.
    Measured checkpoint quality informs subsequent search, while checkpoint materialization provides direct storage measurements.

    \item \textbf{Empirical evaluation of strategy quality and search behavior.}
    We evaluate strategy quality through downstream benchmarks, investigate the search mechanism through matched-seed ablations, and examine allocation structure and reuse through case studies.
\end{itemize}

\section{Related Work}
\label{sec:related}


\subsection{Quantization Strategy Design for Large Language Models}
\label{sec:related_quant}

Post-training quantization (PTQ) reduces the memory and inference cost of
large language models without retraining the full model.
Representative methods mainly improve quantization fidelity under a specified
configuration.
GPTQ uses approximate second-order information for one-shot weight
quantization~\citep{frantar2023gptq}, AWQ protects activation-aware salient
weights through channel scaling~\citep{lin2024awq}, while SmoothQuant
redistributes quantization difficulty between activations and
weights~\citep{xiao2023smoothquant}.
OmniQuant further optimizes clipping and equivalent transformations using a
small calibration set~\citep{shao2024omniquant}.
These methods substantially improve low-bit quantization, but the choice of
quantization configuration itself remains an important design variable.

A complementary line of work therefore treats quantization as a resource
allocation problem.
HAQ automatically searches layer-wise mixed-precision policies using hardware
feedback~\citep{wang2019haq}, while HAWQ-V3 selects mixed-precision
configurations by balancing model perturbation against deployment
constraints~\citep{yao2021hawqv3}.
Such methods primarily vary the precision assigned to different model
components.
AutoQS instead fixes the weight bit-width within each search run and searches
layer- and operator-wise group-size allocations.
It therefore treats \emph{quantization granularity allocation}, rather than
weight precision itself, as the strategy-design variable.


\subsection{LLM-Guided Search in Structured Design Spaces}
\label{sec:related_llmsearch}

Large language models have recently been used as semantic proposal operators
inside iterative search.
Evolution through Large Models (ELM) uses an LLM as a mutation operator in
evolutionary program search~\citep{lehman2022elm}, while FunSearch couples an
LLM with executable evaluation and an evolving program
population~\citep{romeraparedes2024funsearch}.
Evolution of Heuristics (EoH) evolves heuristic ideas and implementations
jointly~\citep{liu2024eoh}, and ReEvo combines evolutionary search with
LLM-based reflection to improve subsequent proposals~\citep{ye2024reevo}.
A common principle across these methods is that evaluated candidates and
search history are fed back to the LLM to guide later generations.

Most existing LLM-driven evolutionary systems operate over comparatively
open-ended spaces such as programs or heuristics.
Quantization strategy search is more constrained: every proposal must preserve
the model structure and fixed precision, assign legal group sizes to all
required operators, and avoid repeatedly evaluating previously visited
policies.
AutoQS therefore uses the LLM as a constrained strategy-search operator rather
than as an unrestricted code generator.
Model sensitivity and search history guide coordinated policy changes, while
policy validation, novelty checking, and nearby refinement keep the search
within a valid structured design space.


\subsection{Budget-Efficient and Diversity-Preserving Search}
\label{sec:related_budget}

High-fidelity evaluation is often the dominant cost in automated design.
Multi-fidelity methods address this problem by evaluating many configurations
cheaply and allocating greater resources only to selected candidates.
Hyperband formalizes this idea through adaptive resource allocation and early
stopping~\citep{li2018hyperband}, while BOHB combines multi-fidelity resource
allocation with model-based search~\citep{falkner2018bohb}.
A similar tension arises in quantization search: inexpensive proxies permit
broad exploration, whereas fresh quantization and downstream evaluation provide
more faithful but substantially more expensive measurements.

Selecting candidates solely by a scalar proxy score can nevertheless
concentrate the search on highly similar configurations.
Quality-diversity methods such as MAP-Elites instead retain high-performing
solutions across distinct regions of a search space~\citep{mouret2015mapelites}.
AutoQS adopts this general diversity-preserving principle without requiring
every candidate to undergo physical evaluation.
Candidate policies are screened using low-cost proxies, structurally distinct
allocations are retained in the search archive, and only a small subset is
promoted to actual GPTQ quantization and downstream evaluation.
This combination of inexpensive screening, diversity preservation, and
selective high-fidelity evaluation allows AutoQS to use a limited physical
evaluation budget while maintaining broad exploration of the policy space.

\section{Method}
\label{sec:method}
\begin{figure}[t]
    \centering
    \includegraphics[width=\textwidth]{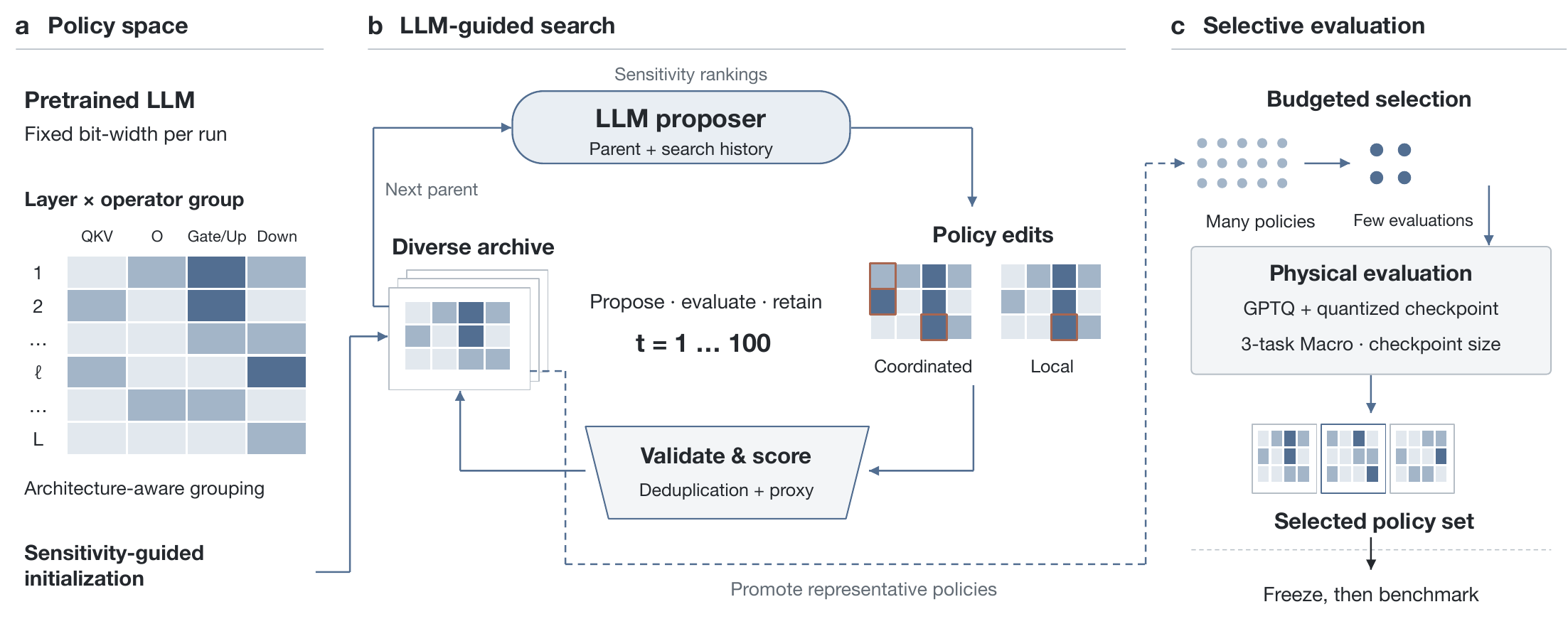}
    \caption{
    Overview of the AutoQS framework.
    AutoQS first defines a structured quantization policy space under a fixed
    weight bit-width, where layer- and operator-wise group sizes are searched.
    An LLM-guided search procedure generates constrained policy modifications,
    which are validated and evaluated using a low-cost proxy while a
    diversity-preserving candidate records maintains previously explored allocation
    patterns.
    Every $K=5$ search iterations, the procedure selects the latest eligible
    candidate in the current five-iteration window for fresh GPTQ quantization.
    This target selection is independent of the proxy-score ordering.
    The resulting checkpoint is evaluated for task accuracy and storage,
    providing measured outcomes for subsequent policy selection.
    }
    \label{fig:autoqs_framework}
\end{figure}

AutoQS searches for group-size allocations while keeping weight precision
fixed. The framework separates inexpensive candidate exploration from
periodic evaluation of actual quantized models. An LLM proposes changes
to an existing allocation. Each proposal is validated and checked for
duplicates, with deterministic novelty repair used to seek an unseen
allocation before proxy evaluation. Evaluated policies and their results
are retained to guide subsequent proposals. Every five iterations, the latest eligible candidate
from the current search window is selected for fresh GPTQ quantization.
The resulting checkpoint provides measured task accuracy and storage cost.
We first define the allocation problem (Sec.~\ref{sec:method_space}), then describe candidate generation (Sec.~\ref{sec:method_search}),
proxy evaluation (Sec.~\ref{sec:method_proxy}), and periodic quantization (Sec.~\ref{sec:method_physical}).

\subsection{Problem Formulation and Feasible Allocations}
\label{sec:method_space}

Consider a pretrained transformer with $L$ layers. The weight bit-width
$b$ is fixed within a search run and is not a search variable.
AutoQS instead assigns group sizes across layers and operator groups.
A group size controls how many weights share quantization parameters.
Smaller groups provide finer granularity but require more scale-related
metadata. Different allocations can therefore change quantization error
and checkpoint size without changing the weight bit-width.

Within each layer, we use four decision groups,
$\mathcal{U}=\{\mathrm{QKV},\mathrm{Out},\mathrm{GateUp},\mathrm{Down}\}$.
QKV contains the query, key, and value projections. Out is the attention
output projection. GateUp contains the gate and up projections of the
feed-forward network, and Down is its down projection.
Coupled operators share a decision to preserve the grouping required
by the target architecture and quantization implementation.
Let $g_{\ell,u}$ denote the group size for group $u$ in layer $\ell$.
A policy contains $4L$ decisions, with feasible allocations defined by
\begin{equation}
    \Pi_b = \left\{
        \pi=(g_{\ell,u})\in\mathcal{G}^{L\times|\mathcal{U}|}
        \;\middle|\; \mathcal{C}_b(\pi)=1
    \right\},
    \qquad \mathcal{G}=\{-1,16,32,64,128\}.
    \label{eq:feasible_space}
\end{equation}
Here $\mathcal{C}_b$ checks the fixed precision, legal group sizes,
model structure, and architecture-dependent grouping constraints.
The value $-1$ denotes a special grouping option of the quantization
implementation rather than a literal negative group size.

The goal is to discover allocations with favorable measured quality
and storage trade-offs under a limited search budget.
Actual quantization is too costly to apply to every proposal, so AutoQS
uses proxy measurements for exploration and periodically evaluates a
smaller set of quantized checkpoints. The proxy and checkpoint measurements
have distinct roles and are not treated as interchangeable evidence.

\subsection{Constrained LLM-Guided Candidate Generation}
\label{sec:method_search}

The allocation space grows exponentially with the number of layers.
Following the iterative modification of existing candidates
in LLM-driven evolutionary search~\citep{novikov2025alphaevolve},
AutoQS explores this space through edits to a parent allocation.
The LLM receives the parent's allocation and evaluation statistics,
representative searched policies, and a summary of search history.
Sensitivity-related statistics are retained as auxiliary Stage-1 evaluation
signals. They are not specified as explicit layer- or operator-level
instructions to the LLM proposer.

Each mutation is expressed as one to four sparse, layer-level
\texttt{SEARCH/REPLACE} blocks. These blocks modify a small number of
complete layer configurations.
Several operator decisions can change together, allowing the LLM to
explore coordinated allocations without regenerating the entire policy.
Candidate alternatives inform the proposal, but all changes are applied
to the currently selected parent. For each modified layer, the proposal's
stated source configuration must match that parent exactly.
A mismatch invalidates the proposal. This prevents contextual examples
or stale allocations from becoming unintended modification targets.

The child policy is then checked against the feasible space in
Eq.~\eqref{eq:feasible_space}. Validation enforces the fixed bit-width,
supported group sizes, and required operator structure.
A canonical representation places the allocation decisions in a fixed
order. Its hash identifies equivalent policies even when their proposal
text differs. Previously evaluated allocations are detected before
another proxy evaluation. A deterministic novelty-repair procedure is applied to seek an unseen
allocation. Any repaired allocation must satisfy the same validity checks
and must be checked again for semantic duplication before evaluation.
This procedure separates the LLM's proposal process from the strict
assignments required by the quantization implementation.

The candidate-generation procedure therefore consists of sparse LLM edits,
validation, and deterministic novelty repair. It does not include an
independent local-refinement or neighborhood-search operator. Novelty repair
addresses repeated assignments; it is not a quality-driven local search.

\subsection{Low-Cost Evaluation and Search Candidate}
\label{sec:method_proxy}

Each valid, unseen candidate first receives a frozen low-cost evaluation.
The evaluator records negative log-likelihood degradation, cross-domain
statistics, sensitivity-related signals, and estimated checkpoint storage.
Negative log-likelihood measures how well the model predicts evaluation
text, with lower values indicating better prediction.
Cross-domain statistics help reveal differences that an aggregate quality
score may obscure. These outputs provide search evidence without requiring
a new deployment checkpoint for every allocation.
Individual proxy statistics are retained separately for diagnostics.
Stage-1 also produces a scalar \texttt{cheap\_proxy\_score}, and
\texttt{combined\_score} is an alias for the same value. Denoting this
scalar by $s_{\mathrm{cheap}}(\pi)$, the implementation records
\begin{equation}
    \texttt{cheap\_proxy\_score}(\pi)
    =\texttt{combined\_score}(\pi)
    =s_{\mathrm{cheap}}(\pi).
    \label{eq:cheap_score_alias}
\end{equation}
This scalar is used in the inexpensive search and candidate state, including
cheap-only candidate ordering. It does not determine the periodic physical
evaluation target. Target selection instead follows the temporal and
eligibility rule in Section~\ref{sec:method_physical}.

For exposition, the allocation-dependent storage estimate can be separated
into a fixed weight-payload term and a quantization-metadata term
\begin{equation}
    \widehat W(\pi;b)
        = W_{\mathrm{weight}}(b)+W_{\mathrm{meta}}(\pi;b).
    \label{eq:storage_proxy}
\end{equation}
The weight term is fixed within a run. Allocation changes primarily affect
the metadata term, including the parameters associated with quantization
groups. The estimate helps compare candidates during search, but the
actual checkpoint may incur additional implementation-dependent overhead.
Measured checkpoint size is therefore obtained separately after quantization.

Maintaining evaluated candidates as context for subsequent
generation is also used in LLM-driven program
search~\citep{romeraparedes2024funsearch}.
AutoQS retains evaluated policies, their scalar proxy scores,
and individual diagnostic statistics in a candidate repository. The candidate records explored allocations and provides earlier
results for subsequent parent selection and proposal generation.
It also preserves access to different allocation patterns rather than
limiting the search context to one incumbent. Representative policies
can include both high-performing and structurally different allocations.
The accompanying visited-policy record supports duplicate detection.

The inexpensive loop thus proceeds from proposal generation to validation,
proxy evaluation, and candidate records update. The recorded evaluation information supports the context
used for later proposals. This supports exploration between actual quantization
attempts. Retaining different policies does not by itself establish their
quality after GPTQ quantization, so the search also performs periodic
checkpoint evaluation.
\begin{algorithm}[t]
\caption{AutoQS Search}
\label{alg:autoqs}
\begin{algorithmic}[1]

\Require pretrained model $M$;
         fixed bit-width $b$;
         feasible policy space $\Pi_b$;
         search budget $T$;
         physical interval $K$
\Ensure search archive $\mathcal{A}$ and physically evaluated policies

\State Initialize a valid policy $\pi_0\in\Pi_b$
\State Initialize archive $\mathcal{A}$ and seen-policy set $\mathcal{H}$

\For{$t=1,\ldots,T$}

    \State $\pi_t^{p}
        \gets
        \Call{SelectParent}{\mathcal{A}}$

    \State $S_t
        \gets
        \Call{BuildContext}{
            \pi_t^{p},
            \mathcal{A},
            \mathcal{H}
        }$

    \State $\Delta_t
        \gets
        \Call{LLMPropose}{
            S_t,
            \mathbf{s}
        }$

    \If{\textbf{not}
        \Call{ParentConsistent}{
            $\Delta_t,
            \pi_t^{p}$
        }}
        \State \textbf{continue}
    \EndIf

    \State $\tilde{\pi}_t
        \gets
        \Call{ApplyModification}{
            \pi_t^{p},
            \Delta_t
        }$

    \If{$\tilde{\pi}_t\notin\Pi_b$}
        \State \textbf{continue}
    \EndIf

    \State $\pi_t
        \gets
        \Call{NoveltyCheckOrRepair}{
            \tilde{\pi}_t,
            \mathcal{H}
        }$

    \State $\widehat{m}_t
        \gets
        \Call{ProxyEvaluate}{
            \pi_t
        }$

    \State $\mathcal{A}
        \gets
        \Call{UpdateArchive}{
            \mathcal{A},
            \pi_t,
            \widehat{m}_t
        }$

    \State $\mathcal{H}
        \gets
        \mathcal{H}
        \cup
        \{\pi_t\}$

    \If{$t \bmod K = 0$}

        \State $\pi_t^{\mathrm{phys}}
            \gets
            \Call{SelectWindowTarget}{
                t,
                K
            }$

        \If{$\pi_t^{\mathrm{phys}}$ does not exist}
            \State \textbf{terminate search}
        \EndIf

        \State $Q_t
            \gets
            \Call{GPTQQuantize}{
                M,
                \pi_t^{\mathrm{phys}},
                b
            }$

        \State $q_t
            \gets
            \Call{Evaluate3Tasks}{
                Q_t
            }$

        \State $c_t
            \gets
            \Call{CheckpointBytes}{
                Q_t
            }$

        \State \Call{RegisterPhysicalResult}{
            $\pi_t^{\mathrm{phys}},
             q_t,
             c_t$
        }

    \EndIf

\EndFor

\State \Return $\mathcal{A}$ and registered high-fidelity policies

\end{algorithmic}
\end{algorithm}
\subsection{Periodic Quantization and Checkpoint Evaluation}
\label{sec:method_physical}

AutoQS reserves an evaluation opportunity every $K$ search iterations.
With $T$ iterations, the number of opportunities is at most
$\lfloor T/K\rfloor$. The reported configuration uses $T=100$ and $K=5$,
allowing at most 20 opportunities. This scheduling limits expensive
quantization attempts while allowing proxy-based exploration between them.
Physical target selection does not rank candidates by quality, storage,
novelty, or the scalar Stage-1 score. An opportunity is distinct from an
actual quantization attempt, and an attempt is distinct from a successful
checkpoint evaluation.

At a boundary iteration $t$ with $t\bmod K=0$, the target comes from the
current window $\mathcal{W}_t=\{t-K+1,\ldots,t\}$. Let $e_j$ indicate
whether iteration $j$ produced an \emph{eligible physical candidate}, with
eligibility assessed at the current boundary. Eligibility requires at least
a valid policy, deployability under the implementation's checks, and a
physical state that is not terminal. It also includes any additional checks
performed by the implementation's target selector. Thus, validity alone is
insufficient. Define the eligible set and selected iteration as
\begin{equation}
    \mathcal{E}_t=\{j\in\mathcal{W}_t\mid e_j=1\},
    \qquad
    j_t^\star=\max\mathcal{E}_t
    \quad\text{if }\mathcal{E}_t\ne\varnothing.
    \label{eq:physical_target}
\end{equation}
If the boundary candidate is eligible, it is selected directly. Otherwise,
the procedure falls back to the latest eligible candidate in the same
window. It does not retrieve a target from an earlier window. The eligibility
check is applied at each boundary even if the boundary iteration produced
an invalid proposal or no new candidate.

If the current window contains no eligible candidate, no target exists.
The stopping rule in the supplied Method terminates the search in this
case. This branch must be checked against the executed controller before
the experimental protocol is finalized.

The selected policy is instantiated through fresh GPTQ quantization
at bit-width $b$~\citep{frantar2023gptq}.
Its checkpoint is evaluated on
ARC-Challenge~\citep{clark2018think},
PIQA~\citep{bisk2020piqa}, and
BoolQ~\citep{clark2019boolq}. These are the three task benchmarks used
for in-search checkpoint assessment. Writing $q_d(\pi)$ for accuracy
on task $d$, the recorded quality score is
\begin{equation}
    Q_{\mathrm{phys}}(\pi)
        = \frac{q_{\mathrm{ARC\text{-}C}}(\pi)
                 +q_{\mathrm{PIQA}}(\pi)
                 +q_{\mathrm{BoolQ}}(\pi)}{3}.
    \label{eq:physical_quality}
\end{equation}
Storage is measured directly as the byte size of the generated checkpoint.
Each successful checkpoint evaluation therefore provides a measured
accuracy and storage observation, linked to its policy identity and
originating search iteration. Failed attempts are recorded as failures
rather than complete accuracy--storage observations. The resulting records
are retained with the search state.

After search, evaluated checkpoints support policy selection for the
broader downstream assessment described in Section~\ref{sec:exp_setup}.
The three in-search benchmarks are part of the development procedure,
so they must be distinguished from any benchmarks reserved for final testing.
The procedure establishes measured accuracy and checkpoint storage.
Runtime latency is not an evaluated objective in the supplied protocol.


\newcommand{\ours}{AutoQS}

\section{Experiments}
\label{sec:experiments}

\subsection{Experimental Setup}
\label{sec:exp_setup}

\paragraph{Models and quantization setting.}
We evaluate \ours{} on Qwen2.5-7B, Qwen2.5-7B-Instruct,
LLaMA-7B, and Llama2-7B, with additional transfer experiments on
Vicuna-7B-v1.5.
For each search run, the weight bit-width is fixed and is not itself a
search variable.
Instead, \ours{} searches heterogeneous group-size allocations across
transformer layers and operator groups, with legal choices
\[
    \{-1,16,32,64,128\}.
\]
Different weight precisions are studied through independent fixed-bit
search runs.

\paragraph{Search and evaluation protocol.}
Unless otherwise specified, each search contains $T=100$ iterations.
Every valid unseen candidate first receives a low-cost Stage-1 evaluation,
while one high-fidelity physical evaluation is scheduled every five
iterations.
The selected policy is freshly materialized with GPTQ
\citep{frantar2023gptq} and evaluated on ARC-Challenge, PIQA, and BoolQ.
We denote the unweighted mean over these three search-time tasks as
\emph{Macro3}.

For post-search evaluation, we follow the seven-task benchmark suite used
in the Qwen2.5-7B evaluation of FBQuant~\citep{liu2025fbquant},
consisting of MMLU, HellaSwag, ARC-Easy, ARC-Challenge, PIQA,
WinoGrande, and BoolQ.
We report their unweighted mean as \emph{Macro7}.
This choice also enables direct comparison with the external quantization
baselines reported by FBQuant.
Macro7 is not fed back into search.
Because ARC-Challenge, PIQA, and BoolQ also appear in the search-time
Macro3 objective, Macro7 is treated as a post-search seven-task summary
rather than a fully held-out benchmark set.

Unless otherwise specified, experiments use one NVIDIA GeForce RTX 3090
GPU with 24\,GB memory, batch size 1 for physical evaluation, and
Qwen-Plus as the default LLM proposer.
Additional protocol, seed, transfer, and compute details are provided in
Appendix~\ref{app:exp_details}.

\subsection{Main Results}
\label{sec:main_results}

\begin{table*}[t]
\caption{
Main 4-bit results on Qwen2.5-7B.
External baseline results are taken from Table~8 of
FBQuant~\citep{liu2025fbquant}.
All external quantized methods use group size 128, whereas \ours{} uses
a searched heterogeneous group-size configuration.
Bold denotes the best result among 4-bit methods.
}
\label{tab:main_qwen25_base_w4}
\centering
\small
\setlength{\tabcolsep}{3.1pt}
\resizebox{\textwidth}{!}{
\begin{tabular}{lcccccccccc}
\toprule
Method & W-bit & Group &
MMLU & HellaSwag & ARC-E & ARC-C & PIQA & WinoGrande & BoolQ &
Avg. $\uparrow$ \\
\midrule

FP16
& 16 & -- &
71.94 & 78.94 & 80.43 & 51.02 &
79.71 & 73.01 & 84.68 & 74.25 \\

\midrule

RTN~\citep{liu2025fbquant}
& 4 & 128 &
71.26 & 75.84 & 77.05 & 48.14 &
77.26 & 70.51 & 80.79 & 71.55 \\

GPTQ~\citep{frantar2023gptq}
& 4 & 128 &
69.78 & 76.68 & 78.21 & 49.58 &
76.15 & 71.08 & 83.16 & 72.09 \\

AWQ~\citep{lin2024awq}
& 4 & 128 &
\textbf{71.31} & 78.26 & 79.17 & 49.15 &
79.60 & 71.67 & 82.91 & 73.15 \\

OmniQuant~\citep{shao2024omniquant}
& 4 & 128 &
71.01 & 78.25 & 79.46 & 50.54 &
78.72 & \textbf{72.29} & 83.61 & 73.41 \\

CALDERA~\citep{saha2024compressing}
& 4 & 128 &
64.13 & 75.61 & 73.50 & 45.54 &
72.83 & 70.03 & 79.83 & 68.78 \\

SVDQuant~\citep{li2024svdquant}
& 4 & 128 &
71.07 & 78.26 & 80.22 & 51.88 &
\textbf{80.47} & 72.14 & 82.91 & 73.85 \\

FBQuant~\citep{liu2025fbquant}
& 4 & 128 &
65.30 & 69.12 & 71.55 & 47.01 &
74.81 & 65.04 & 77.61 & 67.21 \\

\midrule

\ours{}
& 4 & Mixed &
70.43 &
\textbf{78.69} &
\textbf{80.35} &
\textbf{56.57} &
79.54 &
68.43 &
\textbf{85.44} &
\textbf{74.21} \\

\bottomrule
\end{tabular}}
\end{table*}

On Qwen2.5-7B, \ours{} reaches a Macro7 score of 74.21,
exceeding the reported 4-bit baselines in overall average.
The largest task-level gain appears on ARC-Challenge, where the searched
mixed-group policy obtains 56.57 compared with 51.88 for the strongest
external baseline in this table.
These results show that heterogeneous grouping provides an additional
degree of freedom beyond a globally fixed group size while keeping the
weight precision fixed.

The behavior also extends beyond the Qwen2.5-7B base checkpoint.
On Qwen2.5-7B-Instruct, \ours{} reaches 75.16 Macro7, compared with
75.12 for the dense model and 74.46 for uniform-G128 GPTQ.
We further observe that a policy searched on LLaMA-7B can be transferred
to related checkpoints without target-side allocation search, remaining
within 0.15 Macro7 points of target-specific uniform W4-G128 baselines.
Full per-task, transfer, and additional model results are reported in
Appendix~\ref{app:extra_results} and
Appendix~\ref{app:policy_transfer}.

\subsection{Search Effectiveness}
\label{sec:search_effectiveness}

To isolate the contribution of the search mechanism, we compare three
Qwen2.5-7B W4 runs using the same search seed of 1234.
The \emph{Full} configuration uses the Qwen-Plus proposer and incorporates
physical-evaluation feedback into later search decisions.
The \emph{No-feedback} variant retains the same proposer and physical
evaluation schedule but removes physical feedback from subsequent search.
The \emph{Random proposer} variant replaces the LLM proposer with a
deterministic seeded legal random mutation mechanism while preserving the
remaining protocol.

\begin{table*}[t]
\caption{
Matched-seed search-mechanism ablation on Qwen2.5-7B W4.
All runs use search seed 1234, $T=100$ iterations, and one physical
evaluation every five iterations.
``Best Macro3'' denotes the best observed physical candidate under the
predefined search-time criterion.
}
\label{tab:search_mechanism_macro3}
\centering
\small
\setlength{\tabcolsep}{4.0pt}
\resizebox{\textwidth}{!}{
\begin{tabular}{llccccc}
\toprule
Setting & Proposer & Feedback &
Best iter. & Best Macro3 $\uparrow$ & Iter.-100 Macro3 &
Mean policy change (\%) \\
\midrule
Full
& Qwen-Plus
& Yes
& 100
& \textbf{73.8521}
& \textbf{73.8521}
& 26.53 \\

No-feedback
& Qwen-Plus
& No
& 45
& 72.5777
& 72.2177
& 29.23 \\

Random proposer
& Legal random mutation
& Yes
& 15
& 73.4465
& 72.0465
& 11.91 \\
\bottomrule
\end{tabular}}
\end{table*}

Under the matched seed, Full attains the highest best-observed physical
Macro3 and continues improving through iteration 100.
Random proposer reaches its best candidate at iteration 15, whereas
No-feedback peaks at iteration 45 despite making comparably large policy
changes.
Within this controlled comparison, the results support the usefulness of
both structured LLM proposals and physical feedback for guiding the search.

Because Macro3 participates directly in search, this comparison is
interpreted as a search-mechanism ablation rather than an independent
post-search quality ranking.
Complete trajectories, additional proposer and search-seed experiments,
bit-width analysis, and the relationship between search-time Macro3 and
post-search Macro7 are reported in
Appendix~\ref{app:search_dynamics} and
Appendix~\ref{app:extra_results}.

\subsection{Policy Analysis}
\label{sec:policy_analysis}

\begin{figure*}[t]
\centering
\includegraphics[width=0.90\textwidth]{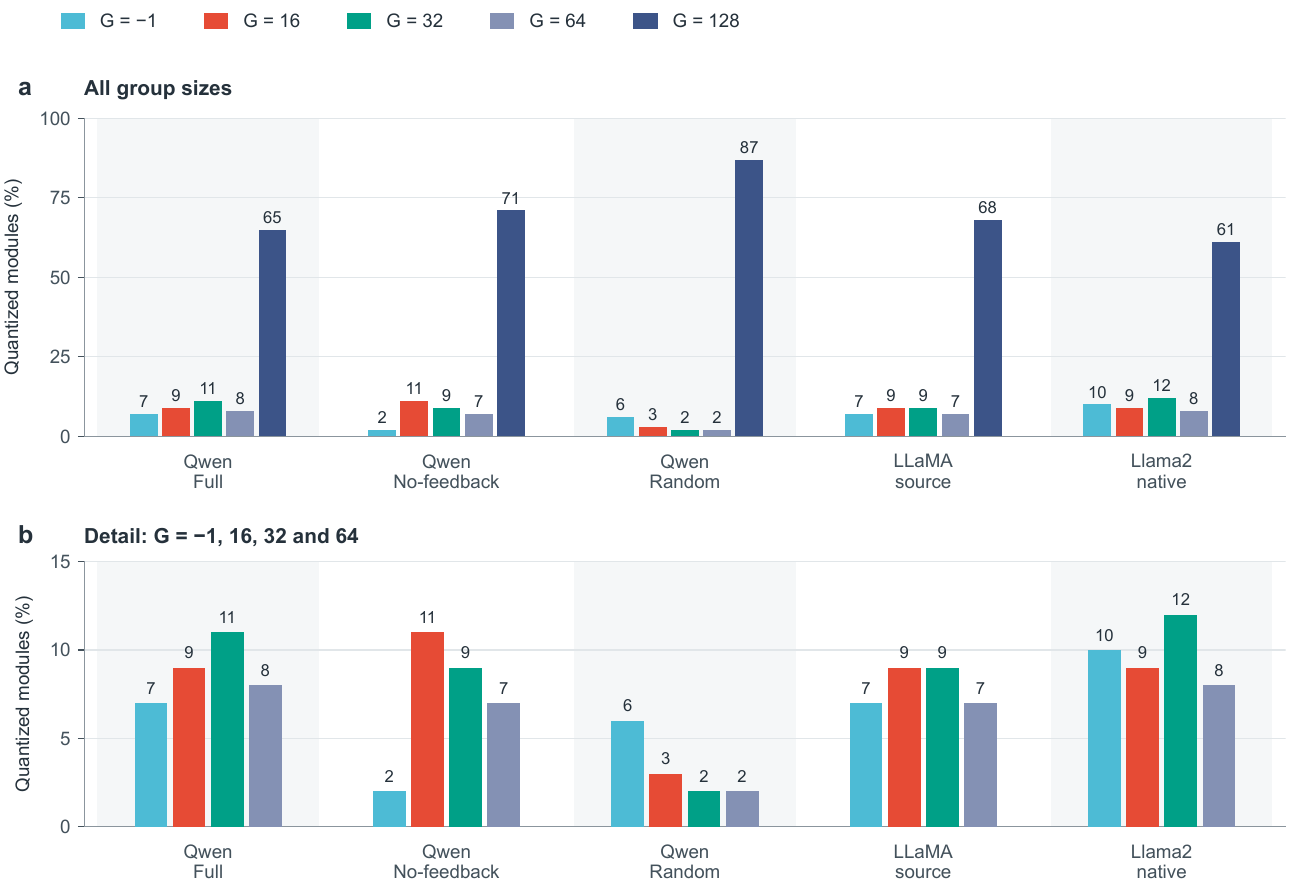}
\caption{
Module-level group-size composition of the analyzed frozen W4 policies.
(a) Full distributions across all five group-size settings.
(b) Detail of the four lower-share settings ($G=-1,16,32,64$).
The three Qwen2.5-7B variants use the same search seed of 1234.
``LLaMA source'' denotes the frozen source policy used for zero-shot
transfer, while ``Llama2 native'' denotes the best observed physical
policy under Macro3 among the 20 physical evaluations of the target-native
Llama2-7B search.
}
\label{fig:selected_policy_composition}
\end{figure*}

The policies discovered by \ours{} are not simply uniform-G128
configurations.
Figure~\ref{fig:selected_policy_composition} summarizes the module-level
group-size composition of several frozen W4 policies.

G128 remains the dominant grouping choice, but a substantial minority of
modules receives alternative group sizes.
For Qwen2.5-7B Full, LLaMA-7B source, and target-native Llama2-7B,
the non-G128 fractions are 28.57\%, 25.00\%, and 30.36\%, respectively.
Thus, \ours{} preserves coarse grouping for most modules while selectively
introducing heterogeneous quantization granularity.

The resulting allocation also depends on the search mechanism.
For example, the matched-seed Random-proposer policy remains substantially
closer to uniform G128 than the Full policy.
Exact module-level counts, pairwise policy differences, and a detailed
comparison of optimization objects are reported in
Appendix~\ref{app:policy_analysis}.

\section{Conclusion}
\label{sec:conclusion}

We presented \ours{}, an automated framework for heterogeneous
quantization-policy design with large language models.
\ours{} fixes the weight bit-width and searches layer- and operator-wise
group-size allocations through LLM-guided proposals, low-cost evaluation,
and periodic physical GPTQ feedback.
Experiments across Qwen and LLaMA-family models show that the searched
policies can match or improve upon uniform-group quantization baselines,
while matched-seed ablations support the roles of structured LLM proposals
and physical feedback.
These results demonstrate the potential of LLM-guided search for
automating quantization strategy design under a limited high-fidelity
evaluation budget.

\subsection*{AI Use Statement}

We used generative AI tools to assist with language editing, translation,
manuscript organization, code development and debugging, and the refinement
of scientific figures and tables. These tools also supported the inspection
of experimental outputs, the summarization of implementation details, and
the presentation of the methodology and experimental analyses.
Generative AI was not used to generate experimental results or fabricate
data. The authors manually reviewed all AI-assisted code, analyses,
figures, numerical results, and manuscript content, and verified them
against the underlying implementation and experimental artifacts.
The authors take full responsibility for the final content of this work,
including all text, claims, code, figures, and results.

\subsection*{Ethics Statement}

This work investigates automated post-training quantization strategy
design for large language models. It does not involve human participants
or personally identifiable information. All experiments use existing
pretrained language models and standard benchmark datasets; no new
datasets were created, and no user data were collected.
We are not aware of ethical concerns specific to the proposed method
beyond the broader considerations associated with the computational use
and deployment of large language models.

\subsection*{Reproducibility Statement}

Section~\ref{sec:method} describes the algorithmic details of \ours{},
including the quantization policy space, candidate-generation procedure,
low-cost evaluation, and periodic physical evaluation protocol.
Section~\ref{sec:exp_setup} presents the main experimental settings,
and Appendix~\ref{app:exp_details} provides further details on the search
protocol, evaluation seeds, transfer setting, and compute budget.
The appendix also includes additional experimental results, search
trajectories, and policy analyses.
To facilitate independent reproduction, we use fixed search budgets,
explicitly specify the group-size choices, and describe the evaluation
protocols used in the reported experiments.

\bibliography{iclr2027_conference}
\bibliographystyle{iclr2027_conference}
\newpage
\appendix
%
%


\section{Additional Experimental Details}
\label{app:exp_details}

\subsection{Search and Evaluation Protocol}

For each \ours{} run, the weight bit-width is fixed throughout the
search and is not itself a search variable.
The search instead assigns heterogeneous group sizes across transformer
layers and operator groups from
\[
    \mathcal{G}=\{-1,16,32,64,128\}.
\]
Different weight precisions are therefore evaluated through independent
fixed-bit runs.

Unless otherwise specified, the formal search contains
$T=100$ iterations.
Every valid unseen candidate first receives the frozen low-cost Stage-1
evaluation.
A high-fidelity physical evaluation is scheduled every five iterations,
yielding at most 20 physical evaluation opportunities per run.

For physical evaluation, the selected allocation is freshly instantiated
using GPTQ and evaluated on ARC-Challenge, PIQA, and BoolQ.
We denote the unweighted mean over these three search-time tasks as
\emph{Macro3}.
The generated quantized checkpoint and its measured storage are retained
together with the physical evaluation result.

After search, frozen checkpoints are evaluated on seven downstream tasks:
MMLU, HellaSwag, ARC-Easy, ARC-Challenge, PIQA, WinoGrande, and BoolQ.
Their unweighted mean is denoted \emph{Macro7}.
These post-search Macro7 results are not fed back into search.

Because ARC-Challenge, PIQA, and BoolQ are already used in the
search-time Macro3 evaluation, the complete seven-task set should not
be interpreted as fully held out.
The remaining four tasks---MMLU, HellaSwag, ARC-Easy, and WinoGrande---
are not used in the Macro3 search objective.

\subsection{Search Seed and Evaluation Seed}

For search-ablation experiments, the seed reported in the tables denotes
the search/evolution seed.
Unless explicitly stated otherwise, post-search Final-7 evaluation uses
random seed 1234.
Search stochasticity and post-search evaluation stochasticity are thus
controlled separately.

This distinction is relevant when comparing multiple searches of the same
model.
For example, differences between Qwen-Plus searches using seeds 1234 and
2026 arise from different search trajectories rather than a change in the
Final-7 evaluator seed.

\subsection{Zero-Shot Quantization-Policy Transfer Protocol}

To evaluate whether a searched allocation policy can be reused beyond its
source checkpoint, we perform zero-shot quantization-policy transfer.

The source allocation is selected using only source-model search-time
Macro3 and is subsequently frozen.
For each target model, we perform no target-side allocation search,
no target-side policy adaptation, and no post-search Final-7 feedback.
The frozen layer- and module-specific group-size assignment is applied
unchanged to the target architecture.

GPTQ calibration and packing are nevertheless performed freshly on the
target weights.
Thus, the transferred object is the quantization allocation policy rather
than the quantized source weights.

\subsection{Search Budget and Retained Accounting}

\begin{table}[t]
\caption{
Compute and evaluation budget of the default 100-iteration \ours{} search.
Only quantities directly supported by the frozen run artifacts are
reported.
}
\label{tab:app_search_cost}
\centering
\small
\begin{tabular}{lc}
\toprule
Metric & Value \\
\midrule
Search iterations & 100 \\
Physical-evaluation & every 5 iterations \\
Maximum physical checkpoints & 20 \\
Physical-evaluation batch size & 1 \\
Default LLM proposer & Qwen-Plus \\
Post-search Final-7 batch size & 1 \\
Post-search Final-7 seed & 1234 \\
LLM token accounting & not retained \\
API cost accounting & not retained \\
\bottomrule
\end{tabular}
\end{table}

The frozen canonical runs do not retain recoverable per-request LLM
token-usage records or API-cost accounting.
We therefore report directly supported physical-evaluation and hardware
quantities rather than retrospectively estimating missing API costs.

\section{Additional Experimental Results}
\label{app:extra_results}

\subsection{Qwen2.5-7B-Instruct Results}

\begin{table*}[t]
\caption{
4-bit results on Qwen2.5-7B-Instruct.
Dense FP16, GPTQ-Int4, AWQ, and \ours{} are evaluated using the same
seven-task post-search protocol.
Bold denotes the best result among 4-bit methods.
}
\label{tab:app_qwen25_instruct_w4}
\centering
\small
\setlength{\tabcolsep}{3.1pt}
\resizebox{\textwidth}{!}{
\begin{tabular}{lcccccccccc}
\toprule
Method & W-bit & Group &
MMLU & HellaSwag & ARC-E & ARC-C & PIQA & WinoGrande & BoolQ &
Avg. $\uparrow$ \\
\midrule
FP16
& 16 & -- &
71.76 & 80.48 & 81.06 & 55.03 &
80.36 & 70.72 & 86.45 & 75.12 \\
\midrule
GPTQ
& 4 & 128 &
70.77 & \textbf{79.73} & 80.13 & 54.35 &
80.25 & \textbf{70.56} & 85.41 & 74.46 \\
AWQ
& 4 & 128 &
\textbf{70.80} & 79.61 & 78.75 & 54.69 &
78.56 & 69.61 & 86.27 & 74.04 \\
\midrule
\ours{}
& 4 & Mixed &
70.71 & 79.68 & \textbf{80.64} & \textbf{57.94} &
\textbf{80.30} & 70.48 & \textbf{86.36} &
\textbf{75.16} \\
\bottomrule
\end{tabular}}
\end{table*}

The searched W4 policy reaches 75.16 Macro7 on
Qwen2.5-7B-Instruct.
It also obtains the strongest reported 4-bit result in this table on
ARC-Easy, ARC-Challenge, PIQA, BoolQ, and the seven-task average.

\subsection{Llama2-7B Baseline Comparison}

\begin{table*}[t]
\caption{
4-bit results on Llama2-7B.
External baseline results are taken from Table~3 of
FBQuant~\citep{liu2025fbquant}.
The external quantized configurations use group size 128.
\ours{}-Transfer applies the frozen allocation searched on the original
LLaMA-7B source model without target-side allocation search or policy
adaptation.
}
\label{tab:app_llama2_w4}
\centering
\small
\setlength{\tabcolsep}{2.9pt}
\resizebox{\textwidth}{!}{
\begin{tabular}{lcccccccccc}
\toprule
Method & W-bit & Group &
MMLU & HellaSwag & ARC-E & ARC-C & PIQA & WinoGrande & BoolQ &
Avg. $\uparrow$ \\
\midrule
FP16
& 16 & -- &
41.84 & 76.00 & 76.35 & 46.25 &
79.11 & 69.06 & 77.71 & 66.62 \\
\midrule
RTN
& 4 & 128 &
39.45 & 74.36 & 74.22 & 42.22 &
77.60 & 66.82 & 74.55 & 64.17 \\
GPTQ
& 4 & 128 &
38.73 & 74.76 & 75.00 & 43.09 &
78.45 & 68.27 & 75.14 & 64.78 \\
AWQ
& 4 & 128 &
40.75 & 75.57 & 75.59 & 45.31 &
78.78 & 68.35 & 78.23 & 66.08 \\
OmniQuant
& 4 & 128 &
41.35 & 74.88 & 75.21 & 44.54 &
78.94 & 68.19 & 77.16 & 65.75 \\
CALDERA
& 4 & 128 &
39.41 & 75.79 & 75.70 & 43.84 &
81.02 & 68.45 & 76.96 & 65.88 \\
SVDQuant
& 4 & 128 &
37.29 & 75.55 & 74.92 & 43.86 &
78.40 & 69.69 & 76.15 & 65.12 \\
FBQuant
& 4 & 128 &
41.42 & 75.30 & 75.84 & 45.37 &
79.11 & 69.64 & 78.50 & 66.45 \\
\midrule
\ours{}-Transfer
& 4 & Mixed &
39.83 & 75.49 & 70.33 & 42.92 &
78.62 & 68.75 & 75.99 & 64.56 \\
\bottomrule
\end{tabular}}
\end{table*}

A separate target-native Llama2-7B search follows the same fixed-W4,
100-iteration protocol.
Among its 20 physical evaluations, the best observed physical candidate
under the predefined Macro3 criterion occurs at physical slot 11,
originating from iteration 55, with Macro3 66.78.
Its post-search evaluation is performed independently and is not used to
change this search-time selection.

\subsection{Cross-Model Policy Transfer}
\label{app:policy_transfer}

\begin{table*}[t]
\caption{
Zero-shot quantization-policy transfer from original LLaMA-7B to related
7B models.
The source policy is selected on LLaMA-7B using source Macro3 only and is
then frozen.
Target models perform no allocation search or policy adaptation;
GPTQ calibration and packing are freshly performed on target weights.
}
\label{tab:zero_shot_transfer}
\centering
\small
\setlength{\tabcolsep}{2.8pt}
\resizebox{\textwidth}{!}{
\begin{tabular}{llcccccccc}
\toprule
Model & Setting &
MMLU & HellaSwag & ARC-E & ARC-C & PIQA & WinoGrande & BoolQ &
Macro7 $\uparrow$ \\
\midrule

LLaMA-7B
& Dense
& 32.17 & 76.20 & 72.85 & 44.80 &
79.16 & 69.93 & 75.17 & 64.33 \\

LLaMA-7B
& Source searched W4
& 27.03 & 73.82 & 71.42 & 42.92 &
79.27 & 70.09 & 74.28 & 62.69 \\

\midrule

Llama2-7B
& Dense (local)
& 41.83 & 76.00 & 74.49 & 46.16 &
79.05 & 68.90 & 77.71 & 66.31 \\

Llama2-7B
& Uniform W4-G128
& 37.96 & 75.71 & 71.17 & 43.60 &
78.51 & 67.48 & 77.00 & 64.49 \\

Llama2-7B
& Zero-shot transfer
& 39.83 & 75.49 & 70.33 & 42.92 &
78.62 & 68.75 & 75.99 & 64.56 \\

\midrule

Vicuna-7B-v1.5
& Dense (local)
& 48.75 & 73.77 & 71.25 & 45.90 &
77.97 & 69.53 & 80.92 & 66.87 \\

Vicuna-7B-v1.5
& Uniform W4-G128
& 47.96 & 73.20 & 70.54 & 44.11 &
77.15 & 69.14 & 80.58 & 66.10 \\

Vicuna-7B-v1.5
& Zero-shot transfer
& 47.91 & 73.21 & 69.99 & 44.03 &
76.88 & 68.59 & 81.07 & 65.95 \\

\bottomrule
\end{tabular}}
\end{table*}

The transferred allocation remains close to a target-specific uniform
W4-G128 baseline on both target models despite using no target-side
allocation search.
On Llama2-7B, zero-shot transfer reaches 64.56 Macro7 compared with
64.49 for uniform W4-G128, a difference of $+0.07$ points.
On Vicuna-7B-v1.5, the transferred policy reaches 65.95 compared with
66.10 for uniform W4-G128, a difference of $-0.15$ points.

These results show that the searched structural allocation can be reused
on related checkpoints without repeating the allocation search, while
remaining close to the corresponding target-specific uniform
quantization baseline \cite{liu2026spark,liu2026elva}.

\subsection{Bit-Width Analysis}

\begin{table*}[t]
\caption{
Bit-width analysis of \ours{} on Qwen2.5-7B.
Each row reports an independently evaluated fixed-bit search.
Within a run, only the heterogeneous group-size allocation is searched\cite{liu2026balance,liu2026instructionasstate}.
}
\label{tab:ablation_bits}
\centering
\small
\setlength{\tabcolsep}{3.1pt}
\resizebox{\textwidth}{!}{
\begin{tabular}{lcccccccc}
\toprule
W-bit &
MMLU & HellaSwag & ARC-E & ARC-C & PIQA & WinoGrande & BoolQ &
Macro7 $\uparrow$ \\
\midrule
2
& 25.02 & 26.18 & 25.13 & 25.26 &
51.69 & 50.20 & 50.09 & 36.22 \\
3
& 58.66 & 70.06 & 65.36 & 45.73 &
73.29 & 58.96 & 74.50 & 63.79 \\
4
& 70.43 & 78.69 & 80.35 & 56.57 &
79.54 & 68.43 & 85.44 & \textbf{74.21} \\
\bottomrule
\end{tabular}}
\end{table*}

Macro7 increases from 36.22 at W2 to 63.79 at W3 and 74.21 at W4.
Thus, heterogeneous allocation search does not fully compensate for the
representational limitations introduced by extremely low weight precision.

\subsection{Additional Proposer, Search-Seed, and Low-Bit Ablations}

\begin{table*}[t]
\caption{
Effect of LLM proposer, search seed, and weight precision on
Qwen2.5-7B.
The reported seed is the search/evolution seed.
Post-search Final-7 evaluation uses seed 1234.
All entries correspond to completed Final-7 evaluations.
}
\label{tab:app_ablation_api_seed_bits}
\centering
\small
\setlength{\tabcolsep}{2.8pt}
\resizebox{\textwidth}{!}{
\begin{tabular}{lcccccccccc}
\toprule
LLM API & W-bit & Search seed &
MMLU & HellaSwag & ARC-E & ARC-C & PIQA & WinoGrande & BoolQ &
Macro7 $\uparrow$ \\
\midrule

Qwen-Plus
& 4 & 2026
& 70.18 & 78.24 & 79.29 & 54.18 &
79.82 & 66.38 & 84.83 & 73.27 \\

Qwen-Plus
& 4 & 1234
& \textbf{70.43} & \textbf{78.69} & \textbf{80.35} &
\textbf{56.57} & 79.54 & \textbf{68.43} &
\textbf{85.44} & \textbf{74.21} \\

DeepSeek
& 4 & 1234
& 70.33 & 78.44 & 79.46 & 54.52 &
\textbf{79.76} & 68.35 & 84.95 & 73.69 \\

\midrule

Qwen-Plus
& 2 & 2026
& 25.49 & 26.12 & 24.87 & \textbf{27.99} &
51.96 & 49.57 & 48.78 & 36.40 \\

DeepSeek
& 2 & 1234
& \textbf{25.86} & \textbf{26.34} & \textbf{25.00} &
27.30 & \textbf{53.10} & \textbf{50.51} &
\textbf{49.97} & \textbf{36.87} \\

\bottomrule
\end{tabular}}
\end{table*}

At W4, Qwen-Plus with search seed 1234 obtains 74.21 Macro7,
compared with 73.27 for search seed 2026 and 73.69 for the DeepSeek
proposer at search seed 1234.
Because the post-search evaluator uses the same evaluation seed, the
difference between the two Qwen-Plus runs reflects variation in the
search trajectory rather than a change in evaluation randomness.

The W2 results are included as additional evidence of the severe quality
loss in the extremely low-bit regime \cite{fu2026hero,liu2026autoprune,liu2026codesemantics}.

\section{Search Dynamics and Diagnostics}
\label{app:search_dynamics}

\subsection{Matched-Seed Search Trajectories}

We further inspect how physical performance and the module-level policy
evolve in the matched-seed Qwen2.5-7B W4 experiments.
Because a physical evaluation is scheduled every five search iterations,
each 100-iteration run contains up to 20 physical trajectory points.

\begin{figure*}[t]
\centering
\includegraphics[width=\textwidth]
{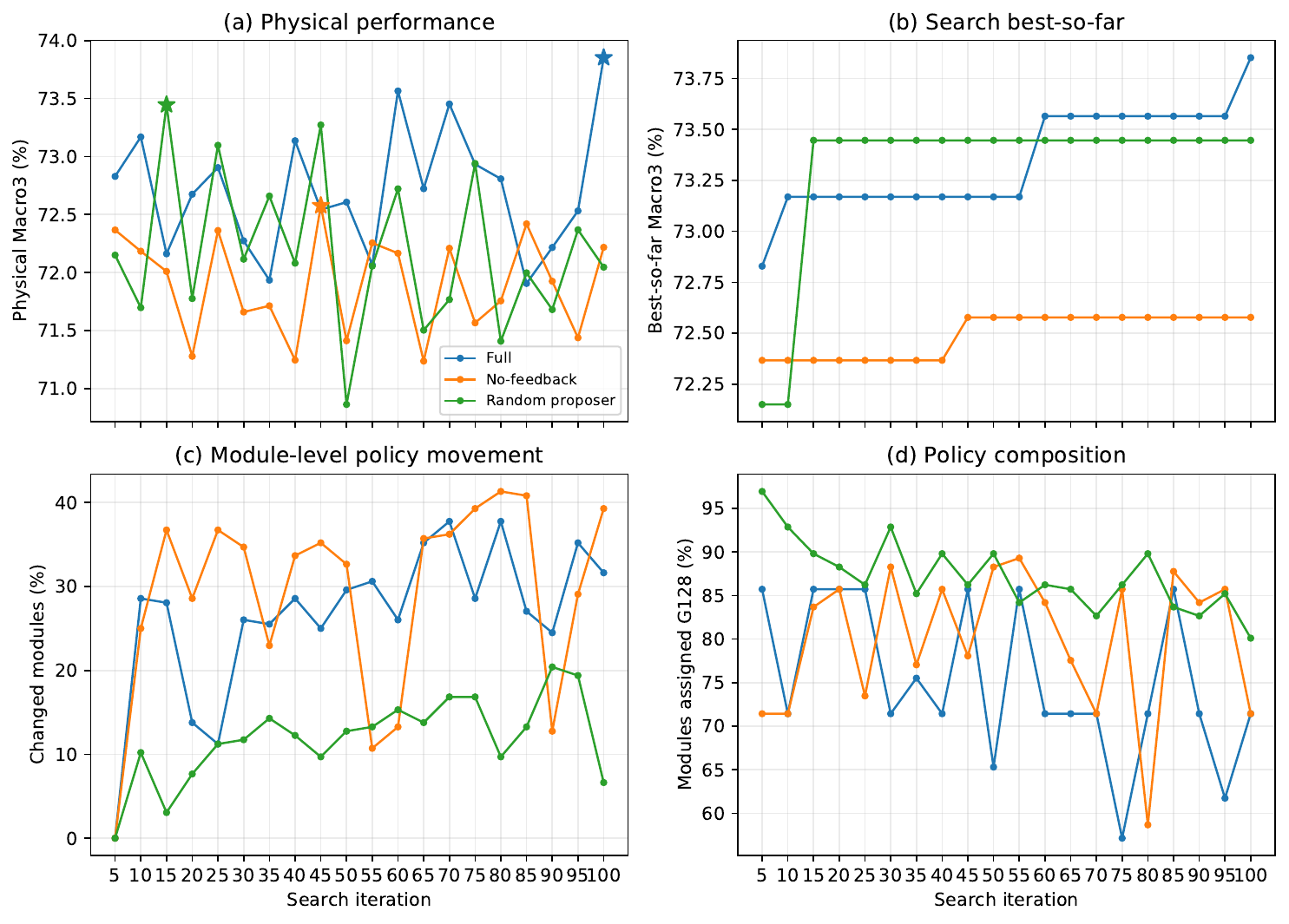}
\caption{
Matched-seed evolutionary search dynamics on Qwen2.5-7B W4.
All three runs use search seed 1234, 100 search iterations, and one
physical evaluation every five iterations.
(a) Physical Macro3 at each physical boundary; stars mark the best
observed physical candidate under the predefined Macro3 criterion.
(b) Best-so-far Macro3.
(c) Fraction of module-level group-size assignments changed relative to
the preceding physical policy; the first point is set to zero by
convention.
(d) Fraction of modules assigned group size 128.
}
\label{fig:qwen_matched_seed1234_dynamics}
\end{figure*}

The three mechanisms exhibit different temporal behavior.
Full continues to improve late in the run and reaches its best physical
Macro3 of 73.85 at iteration 100.
Random proposer reaches its best value of 73.45 at iteration 15 and
does not improve its best-so-far score thereafter.
No-feedback peaks at 72.58 at iteration 45.

The difference is not explained solely by the magnitude of movement in
policy space.
Between consecutive physical evaluations, Full and No-feedback change
26.53\% and 29.23\% of module-level group assignments on average,
respectively, whereas Random proposer changes 11.91\%.
Thus, No-feedback still performs substantial policy updates, but those
updates do not yield the same late-stage improvement in the physical
search objective.

Random proposer also remains closer to a G128-dominated region:
its G128 fraction ranges from 80.10--96.94\%, whereas Full ranges from
57.14--85.71\%.

\subsection{Post-Search Evaluation of Intermediate Checkpoints}

The main text reports only the final selected search result.
To characterize how post-search quality behaves along a single trajectory,
we independently evaluate selected frozen intermediate checkpoints.

\begin{table*}[t]
\caption{
Post-search Final-7 performance of selected frozen physical checkpoints
from the same 100-iteration Qwen2.5-7B W4 trajectory.
These checkpoints are intermediate states of one search and should not be
interpreted as independent truncated-budget experiments.
}
\label{tab:app_qwen_search_convergence}
\centering
\small
\setlength{\tabcolsep}{3.0pt}
\resizebox{\textwidth}{!}{
\begin{tabular}{ccccccccc}
\toprule
Iteration &
MMLU & HellaSwag & ARC-E & ARC-C & PIQA & WinoGrande & BoolQ &
Macro7 $\uparrow$ \\
\midrule

10
& 70.48 & 78.37 & 80.60 & 55.80 &
79.27 & 68.03 & 84.43 & 73.86 \\

25
& 70.10 & 77.93 & 80.56 & 54.95 &
78.29 & 67.56 & 85.47 & 73.55 \\

60
& 69.98 & 78.52 & \textbf{80.89} & 56.23 &
\textbf{80.03} & 67.01 & 84.43 & 73.87 \\

70
& 69.91 & 78.43 & 80.51 & 56.06 &
79.65 & 67.25 & 84.65 & 73.78 \\

100
& \textbf{70.43} & \textbf{78.69} & 80.35 &
\textbf{56.57} & 79.54 & \textbf{68.43} &
85.44 & \textbf{74.21} \\

\bottomrule
\end{tabular}}
\end{table*}

The iteration-100 checkpoint obtains the highest Macro7 among these
evaluated trajectory points, but the trajectory is not monotonic.
Macro7 decreases from 73.86 at iteration 10 to 73.55 at iteration 25,
rises to 73.87 at iteration 60, declines to 73.78 at iteration 70,
and finally reaches 74.21 at iteration 100.

These checkpoints are therefore used only to characterize the trajectory,
rather than as evidence for independent reduced-budget searches.

\subsection{Search-Time Macro3 versus Post-Search Macro7}

We additionally examine whether search-time Macro3 provides the same
candidate ordering as post-search Macro7.

\begin{table}[t]
\caption{
Search-time Macro3 and post-search Macro7 for eight frozen physical
candidates from the LLaMA-7B W4 trajectory.
The search rank is determined exclusively by Macro3, whereas the Macro7
rank is computed post hoc for analysis.
}
\label{tab:app_proxy_final7_ranking}
\centering
\small
\setlength{\tabcolsep}{3.5pt}
\begin{tabular}{ccrrrr}
\toprule
Slot & Iter. & Macro3 & Macro7 & Search rank & Macro7 rank \\
\midrule
4  & 20  & 64.9683 & \textbf{62.9818} & 8 & 1 \\
6  & 30  & 65.3866 & 62.8990 & 2 & 2 \\
9  & 45  & 65.0425 & 62.6471 & 6 & 7 \\
10 & 50  & 64.9990 & 62.5527 & 7 & 8 \\
17 & 85  & 65.3613 & 62.7483 & 3 & 4 \\
18 & 90  & 65.2977 & 62.8870 & 4 & 3 \\
19 & 95  & 65.1478 & 62.6976 & 5 & 5 \\
20 & 100 & \textbf{65.4901} & 62.6904 & 1 & 6 \\
\bottomrule
\end{tabular}
\end{table}

Across these eight candidates, Pearson correlation is $r=0.135$ and
Spearman rank correlation is $\rho=0.071$ ($n=8$).
The sample size is too small to support a claim of statistical
independence.
It nevertheless shows that Macro3 is not an exact surrogate for Macro7:
the iteration-20 candidate ranks eighth according to Macro3 but first
according to Macro7, whereas iteration 100 ranks first by Macro3 but sixth
by Macro7.

The iteration-100 policy remains the source policy used in the transfer
experiment because it was selected under the predefined search-time
criterion.
Selecting a different policy after observing Macro7 would constitute
post-hoc selection using the post-search evaluation.

\section{Quantization Policy Analysis}
\label{app:policy_analysis}

\subsection{Complete Module-Level Policy Statistics}

The main paper visualizes the group-size composition of representative
frozen policies.
Table~\ref{tab:app_policy_structure} provides the corresponding exact
module-level counts.

\begin{table*}[t]
\caption{
Module-level group-size structure of frozen W4 policies.
The three Qwen2.5-7B search-mechanism rows use the same search seed
of 1234.
The LLaMA-7B row is the frozen source policy used in the transfer
experiment.
The Llama2-7B row is the best observed physical policy under Macro3
among the 20 physical evaluations of the target-native search.
}
\label{tab:app_policy_structure}
\centering
\small
\setlength{\tabcolsep}{3.5pt}
\resizebox{\textwidth}{!}{
\begin{tabular}{llrrrrrrr}
\toprule
Model & Setting & Modules &
$G=-1$ & $G=16$ & $G=32$ & $G=64$ & $G=128$ &
Non-G128 (\%) \\
\midrule

Qwen2.5-7B
& Full, seed 1234
& 196 & 12 & 14 & 15 & 15 & 140 & 28.57 \\

Qwen2.5-7B
& No-feedback, seed 1234
& 196 & 1 & 14 & 13 & 15 & 153 & 21.94 \\

Qwen2.5-7B
& Random proposer, seed 1234
& 196 & 8 & 5 & 4 & 3 & 176 & 10.20 \\

\midrule

LLaMA-7B
& Source searched W4
& 224 & 11 & 15 & 15 & 15 & 168 & 25.00 \\

Llama2-7B
& Target-native best physical
& 224 & 15 & 11 & 24 & 18 & 156 & 30.36 \\

\bottomrule
\end{tabular}}
\end{table*}

G128 remains the dominant group-size choice, while non-G128 assignments
occupy 28.57\% of the Qwen2.5-7B Full policy, 25.00\% of the LLaMA-7B
source policy, and 30.36\% of the target-native Llama2-7B policy.
The matched-seed Random-proposer policy is substantially closer to
uniform G128, assigning G128 to 176 of 196 modules.

\subsection{Pairwise Differences between Matched-Seed Policies}

\begin{table}[t]
\caption{
Pairwise module-level disagreement between matched-seed
Qwen2.5-7B W4 policies.
}
\label{tab:app_policy_disagreement}
\centering
\small
\begin{tabular}{lrr}
\toprule
Policy pair & Different modules & Fraction (\%) \\
\midrule
Full vs.\ No-feedback & 86 / 196 & 43.88 \\
Full vs.\ Random proposer & 63 / 196 & 32.14 \\
No-feedback vs.\ Random proposer & 63 / 196 & 32.14 \\
\bottomrule
\end{tabular}
\end{table}

The matched-seed variants therefore do not merely perturb a single common
final allocation.
They reach materially different module-level policies despite sharing the
same model, bit-width, legal group-size set, search seed, and physical
evaluation cadence.

\subsection{What \ours{} Optimizes Compared with Existing Quantizers}
\label{app:optimization_object}

The comparison in the main paper separates the
\emph{quantization algorithm} from the \emph{allocation policy}.
In the reported W4 experiments, the external baseline configurations use
a globally fixed group size of 128.
In contrast, \ours{} keeps weight precision fixed while exposing
module-level group size as a discrete search variable.

The legal allocation choices are
\[
    \mathcal{G}=\{-1,16,32,64,128\}.
\]

\begin{table*}[t]
\caption{
Comparison of the optimization dimensions represented by the quantization
baselines and \ours{}.
The ``Grouping in our tables'' column refers specifically to the
configurations evaluated in this paper.
}
\label{tab:app_method_strategy_comparison}
\centering
\small
\setlength{\tabcolsep}{3.2pt}
\resizebox{\textwidth}{!}{
\begin{tabular}{llllll}
\toprule
Method &
Primary optimization object &
Typical adaptation level &
Uses calibration signal &
Grouping in our tables &
Relationship to \ours{} \\
\midrule

RTN &
Direct rounding to the target grid &
Weight / quantization group &
No learned optimization &
Fixed G128 &
Reference fixed policy \\

GPTQ &
Sequential one-shot weight quantization using approximate second-order
information &
Weight / block &
Yes &
Fixed G128 &
Physical quantizer used after policy search \\

AWQ &
Activation-aware protection through channel scaling &
Channel &
Yes, activation statistics &
Fixed G128 &
Orthogonal to module-level allocation \\

OmniQuant &
Learnable clipping and equivalent transformations &
Block / channel &
Yes, differentiable calibration &
Fixed G128 &
Optimizes quantization parameters rather than allocation topology \\

CALDERA &
Low-precision plus low-rank matrix decomposition &
Layer / matrix &
Yes &
Fixed G128 in reported comparison &
Changes the weight representation itself \\

SVDQuant &
SVD/low-rank-oriented baseline as reported by FBQuant &
Method-specific &
Method-specific &
Fixed G128 in reported comparison &
No module-level group policy inferred here \\

FBQuant&
Feedback-inspired reconstructed weights with auxiliary sub-branches &
Layer / representation &
Yes &
Fixed G128 &
Feedback acts inside weight reconstruction \\

\midrule

\ours{} &
Discrete assignment of group size to each quantized module &
Module &
Cheap proxy + scheduled physical Macro3 &
Mixed $\{-1,16,32,64,128\}$ &
Searches the configuration consumed by the physical quantizer \\

\bottomrule
\end{tabular}}
\end{table*}

GPTQ and \ours{} therefore operate at different levels of the overall
quantization problem.
GPTQ determines how weights are quantized under a specified configuration,
whereas \ours{} searches which grouping configuration is supplied to the
physical quantizer.

Likewise, AWQ and OmniQuant adapt scaling, clipping, or related
quantization parameters, while CALDERA modifies the weight representation.
The additional degree of freedom explored by \ours{} is the structural
allocation of grouping granularity across modules.

The meaning of ``feedback'' also differs across methods.
In \ours{}, physical feedback refers to downstream Macro3 measurements
from freshly materialized quantized checkpoints that can influence later
search decisions.
This occurs at the allocation-search level rather than inside the
weight-reconstruction procedure.

\section{Automation and Computational Considerations}
\label{app:automation}

\subsection{Size of the Allocation Space}

The motivation for automated search is most apparent from the size of the
allocation problem.
With five legal group-size choices per quantized module, a model with
$M$ quantized modules admits
\[
    |\mathcal{G}|^M = 5^M
\]
possible module-level assignments before any additional search constraints
are considered.

For Qwen2.5-7B, the compiled policy covers 196 modules, giving
\[
    5^{196} \approx 10^{137.0}
\]
possible assignments.
For the 224-module LLaMA-7B and Llama2-7B policies, the corresponding
space is
\[
    5^{224} \approx 10^{156.6}.
\]

Exhaustively evaluating these spaces is infeasible.
Likewise, manually constructing a separate group-size choice for every
module does not scale with model size.
\ours{} instead performs proposal generation, legality checking,
low-cost evaluation, periodic physical evaluation, and search-state
updates automatically under a fixed protocol.

\subsection{Machine Time versus Human Design Time}
\label{app:time_analysis}

It is important to distinguish machine time from model-specific policy
design effort.

\paragraph{Machine time.}
One-shot and calibration-based quantizers may require substantially less
machine time than a 100-iteration search.
For example, GPTQ reports quantizing a 175B-parameter model in
approximately four GPU-hours under its hardware setting, while OmniQuant reports approximately
1--16 hours for LLaMA-2 models from 7B to 70B on a single A100-40GB GPU
using 128 calibration samples.

These values provide context but are not normalized runtime comparisons
with our RTX 3090 experiments because model scale, hardware,
calibration workloads, and optimization objectives differ.

\paragraph{Human design time.}
The principal automation advantage of \ours{} is not a demonstrated
universal reduction in raw GPU time.
Instead, it automates repeated model-specific decisions over the
heterogeneous allocation policy.

A manually engineered heterogeneous allocation would require repeated
decisions about which modules should be modified, which group size each
selected module should use, and whether the resulting policy should be
revised after quantization and evaluation.
In \ours{}, these decisions are executed by the fixed search procedure
after the run is launched.

\begin{table*}[t]
\caption{
Conceptual comparison of manual heterogeneous-policy tuning and the
automated \ours{} workflow.
No numerical human-hours speedup is claimed because manual research time
was not measured under a controlled protocol.
}
\label{tab:app_human_time}
\centering
\small
\setlength{\tabcolsep}{4.5pt}
\begin{tabular}{
p{0.27\textwidth}
p{0.31\textwidth}
p{0.31\textwidth}}
\toprule
Stage &
Manual heterogeneous-policy tuning &
\ours{} \\
\midrule

Choose modules to modify &
Researcher manually selects target modules &
Candidate generation and mutation are performed automatically \\

Choose group size &
Researcher chooses a value for each target &
Values are selected from the legal set under the search protocol \\

Validate legality &
Researcher or ad-hoc scripts verify compatibility &
Semantic and structural checks reject invalid candidates automatically \\

Low-cost screening &
Researcher designs and runs exploratory tests &
Stage-1 evaluation is applied to every valid unseen candidate \\

High-fidelity test &
Researcher manually launches quantization and evaluation &
Physical GPTQ + Macro3 is scheduled automatically \\

Use previous result &
Researcher inspects results and chooses the next modification &
Physical feedback can influence later search decisions \\

Repeat &
Requires another manual design/evaluation cycle &
Search continues automatically to the fixed iteration budget \\

\bottomrule
\end{tabular}
\end{table*}

Accordingly, the strongest time-related claim supported by the current
experiments is an \emph{automation advantage} rather than a universal
machine-time speedup.

\subsection{Limitations of the Strategy and Time Comparison}
\label{app:comparison_limitations}

The preceding analysis has several limitations.

First, the fixed-G128 descriptions of external baselines refer only to the
configurations evaluated in the corresponding result tables.
They should not be interpreted as claims that those methods could never
be combined with other grouping configurations.

Second, we do not report a numerical human-hours speedup because no
controlled record exists for the amount of researcher time required to
manually design an equally strong heterogeneous allocation.
A retrospective numerical estimate would therefore be speculative.

Third, the current module-level structural analysis covers Qwen2.5-7B,
LLaMA-7B, Llama2-7B, and the matched-seed Qwen ablations.
Qwen2.5-7B-Instruct is included in the accuracy experiments, but its exact
compiled module-level allocation is not included in the structural
comparison because the corresponding frozen policy artifact has not yet
been recovered.

These limitations do not change the central observation that \ours{}
turns a very large model-specific discrete allocation problem into a
reproducible automated search procedure, while post-search Final-7
evaluation remains outside the search loop.

\end{document}